\def\year{2022}\relax
\documentclass[letterpaper]{article} 
\usepackage{aaai22}  
\usepackage{times}  
\usepackage{helvet}  
\usepackage{courier}  
\usepackage[hyphens]{url}  
\usepackage{graphicx} 
\usepackage{natbib}  
\usepackage{caption} 
\DeclareCaptionStyle{ruled}{labelfont=normalfont,labelsep=colon,strut=off} 
\usepackage{algorithm}
\usepackage{algorithmic}

\usepackage{microtype}
\usepackage{graphicx}
\usepackage{booktabs} 
\usepackage[short]{optidef}
\usepackage{amsmath}
\usepackage{booktabs}
\usepackage{multirow}
\usepackage{subfig}
\usepackage{xr}
\usepackage{nameref,zref-xr}
\usepackage{tikz}

\DeclareMathOperator{\tr}{tr}

\DeclareMathOperator*{\argmin}{argmin}
\DeclarePairedDelimiter{\norm}{\lVert}{\rVert} 

\title{DARTS-PRIME: Regularization and Scheduling Improve Constrained Optimization in Differentiable NAS}

    \author{
        Kaitlin Maile,\textsuperscript{\rm 1} 
        Erwan Lecarpentier,\textsuperscript{\rm 1,2} 
        Herv\'e Luga,\textsuperscript{\rm 1} 
        Dennis G. Wilson\textsuperscript{\rm 3}
    }
    \affiliations{
        \textsuperscript{\rm 1}IRIT, University of Toulouse, Toulouse, France\\
        \textsuperscript{\rm 2}IRT Saint-Exupery, Toulouse, France\\
        \textsuperscript{\rm 3}ISAE-SUPAERO, University of Toulouse, Toulouse, France\\
        kaitlin.maile@irit.fr
    }
\begin{document}

\maketitle
\begin{abstract} 
Differentiable Architecture Search (DARTS) is a recent neural architecture search (NAS) method based on a differentiable relaxation. Due to its success, numerous variants analyzing and improving parts of the DARTS framework have recently been proposed. By considering the problem as a constrained bilevel optimization, we present and analyze DARTS-PRIME, a variant including improvements to architectural weight update scheduling and regularization towards discretization. We propose a dynamic schedule based on per-minibatch network information to make architecture updates more informed, as well as proximity regularization to promote well-separated discretization. Our results in multiple domains show that DARTS-PRIME improves both performance and reliability, comparable to state-of-the-art in differentiable NAS. \footnote{Our code is available anonymously here: https://anonymous.4open.science/r/DARTS-PRIME}
\end{abstract}

\section{Introduction}

Since their inception, neural networks have progressed from being completely hand-designed to being more and more automated, allowing for larger size, more complex tasks, and better performance. The introduction of back-propagation to optimize parameters within an architecture was a major step towards automation \cite{rumelhart1986}, but until recently, network architectures were still hand-designed, limiting innovation. Neural architecture search has emerged as the path towards automating the structural design of networks. Initial methods used evolutionary algorithms \cite{real2019regularized} or reinforcement learning \cite{zoph2017neural} to search for networks from building blocks of operators. By relaxing the search space over operators to be continuous, the original Differentiable Architecture Search (DARTS) algorithm was the first to use gradient descent for searching across network architectures \cite{liu2018darts}. This one-shot search builds a supernetwork with every possible operator at every possible edge connecting activation states within the network, using trainable structural weights across operators within an edge to determine which edges and operators will be used in the final architecture.

While DARTS was able to drastically reduce search time while finding high-performing networks, it suffers from several drawbacks. Among them, contributions pointed out the lack of early search decisions \cite{li2020pd,chen2019progressive, wang2020rethinking}, failure to approach the final target constraint \cite{chu2020fair}, performance collapse due to skip-connections \cite{chu2020darts,zhou2020theory,zela2020understanding}, and inefficiency due to unconstrained search \cite{yao2020efficient}. 

In this work, we posit that difficulties arise in DARTS due to a lack of informed schedule and regularization befitting a constrained bilevel optimization problem. We formulate our contributions in a new augmentation called \textbf{DARTS-PRIME} (\textbf{D}ifferentiable \textbf{Ar}chi\textbf{t}ecture \textbf{S}earch with \textbf{P}roximity \textbf{R}egularization and Fisher \textbf{I}nfor\textbf{me}d Schedule). Specifically, DARTS-PRIME is a combination of two additions to DARTS:
\begin{enumerate}
\item a dynamic update schedule, yielding less frequent architecture updates with respect to weight updates, using gradient information to guide the update schedule, and
\item a novel proximity regularization towards the desired discretized architecture, which promotes clearer distinctions between operator and edge choices at the time of discretization.
\end{enumerate}

We analyze these two additions together as DARTS-PRIME, as well as each independently in ablation studies, over standard benchmarks and search spaces with CIFAR-10, CIFAR-100, and Penn TreeBank. We find that these additions independently improve DARTS and combine as DARTS-PRIME to make significant accuracy improvements over the original DARTS method and more recently published variants across all three datasets.

\section{Background and Related Work}

In this section we offer an explanation of the DARTS algorithm formulated as a constrained bilevel optimization problem, focusing on the issues which arise due to this complex search. We also cover related variants of DARTS which propose various solutions to the difficulties arising from this constrained bilevel problem.

DARTS sets up a supergraph neural network that contains all possible options in all possible connections within a framework architecture derived from that of other NAS works \cite{zoph2017neural,real2019regularized}. For image classification tasks, this supernetwork consists of repeated cells of two different types, normal and reduce. Each cell of the same type shares the same architecture. This formulation allows for a shallower network with fewer normal cells to be searched over and a deeper one to be used for evaluation, such that both networks fit maximally on a standard GPU. The structure within each cell is a fully-connected directed acyclic graph, receiving input from the two previous cells and using a fixed number of intermediate activation states that are concatenated to produce that cell's output. Each of these edges contains all operators, each of which is a small stack of layers such as a skip-connection or a nonlinearity followed by convolution followed by batchnorm. In order to match the search space to that of the preceding NAS works, the search objective is to find exactly two operators from unique sources, either one of the two previous cells or one of the previous activation states within the cell, to go to each activation state. DARTS accomplishes this by keeping the two strongest operators, as measured by trained continuous architecture weights, from unique sources for each state within the cell. 

The DARTS search can be formalized as follows. Let $\alpha_{ijp}$ be the architecture weight for operator $p$ from node $j$ to node $i$, so $\alpha_{ij}$ represents the set of architecture weights for all operators from node $j$ to node $i$ and $\alpha_{i}$ represents the set of architecture weights for all operators from all nodes to node $i$. The discretization for the DARTS search space may be stated as the intersection of the following sets:
\begin{align}
  S =& \bigcap\limits_{i} S_i^1 \cap S_i^2 \cap S_i^3 \label{eq:Sset} \\
  S_i^1 =& \{\alpha_i | \forall j \text{ card}(\alpha_{ij}) \le 1\} \label{eq:S1}\\
  S_i^2 =& \{\alpha_i | \text{card}(\alpha_i) = 2\}  \label{eq:S2}\\
  S_i^3 =& \{\alpha_i | \alpha_{ijp} \in \{0,1\}\} \label{eq:S3},
\end{align}
where $S_i^1$ states that there is at most one active operator coming in to each intermediate state from any given source node, $S_i^2$ states that there are exactly two total active operators coming in to each intermediate state, and $S_i^3$ states that all $\alpha_{ijp}$ values are binary, so all operators are either active or inactive. Thus, $S$ contains the encoding for all possible discretized architectures $\alpha$ in the search space of DARTS.

\begin{figure}[ht!]
\begin{minipage}{.9\linewidth}
\centering
\subfloat[Normal cell $\alpha$ over search training]{\includegraphics[trim=5 0 0 20, clip,width = 2.7in]{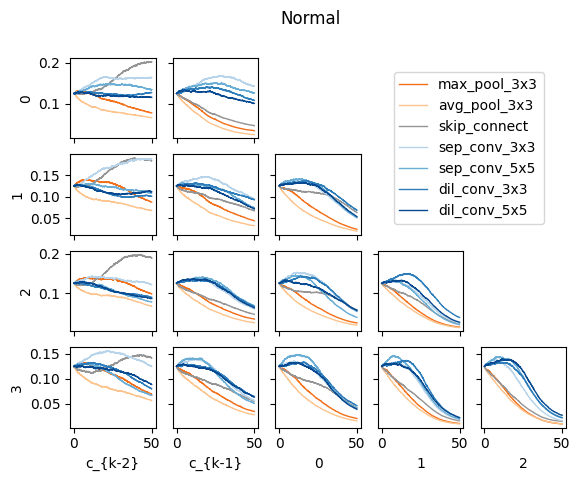}\label{fig:dartsNH2}} \end{minipage}\par
\begin{minipage}{.9\linewidth}
\centering
\subfloat[Reduce cell $\alpha$ over search training]{\includegraphics[trim=5 0 0 20, clip,width = 2.7in]{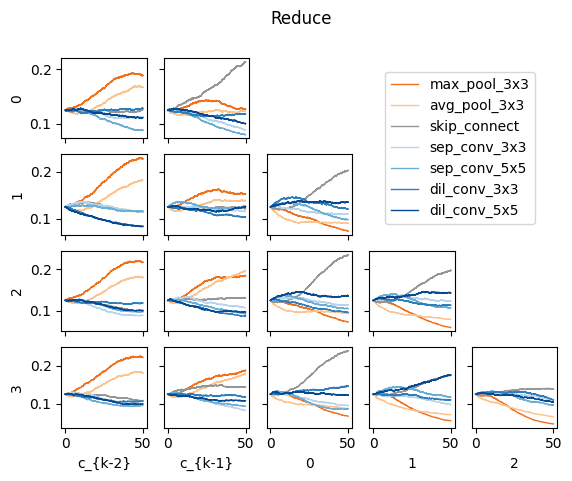}\label{fig:dartsRH2}}
\end{minipage}\par
\begin{minipage}{.395\linewidth}
\centering
\subfloat[Final normal cell]{\makebox[\linewidth][c]{\includegraphics[width = \linewidth]{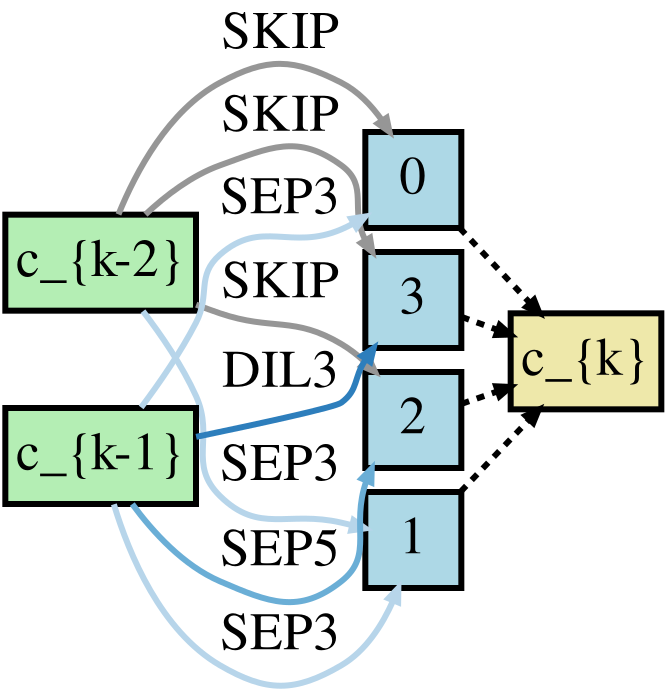}\label{fig:dartsNG2}}}
 \end{minipage}
\begin{minipage}{.595\linewidth}
\centering
\subfloat[Final reduce cell]{\makebox[\linewidth][c]{\includegraphics[width = \linewidth]{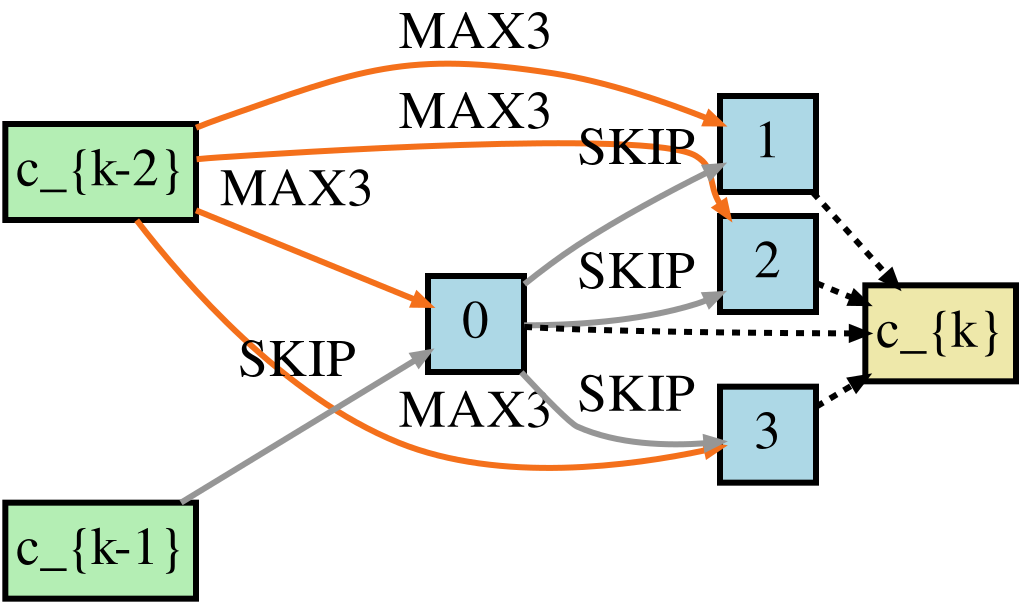}\label{fig:dartsRG2}}}
 \end{minipage}%
 \small
\caption{Training progression and final results of our highest performing trial of DARTS on CIFAR-10. Each subplot in (a) and (b) represents an edge in the supernetwork cell: each column is the source of the edge, and each row is the destination of the edge. The softmaxed $\alpha$ for each operator is plotted over the search training. To derive the architecture shown in (c) and (d) respectively, the two largest activated $\alpha$ values from unique columns are selected in each row at the end of search. The "none" operator is not shown, as it cannot be selected at discretization.}
\label{fig:darts}
\end{figure}

During the DARTS search, $S$ is relaxed so that $\alpha$ is not constrained. A softmax operation is applied over continuous $\alpha$ across operators in an edge as an activation. During a forward pass, the output of each edge is computed as the weighted sum of each operator on that edge multiplied by its activated architecture weight; then, the output of each state is computed as the sum of all incoming edges; finally, the output of the entire cell is computed as the channel-wise concatenation of each intermediate state. After searching for a set number of epochs, the softmaxed $\alpha$ is projected onto $S$ to derive the final architecture, as shown in Figure \ref{fig:darts}.

In order to train the search supernetwork with architecture weights $\alpha$ and network parameters $w$, DARTS optimizes the following objective:
\begin{subequations}
\label{eq:bilevel}
 \begin{align}
 \min_\alpha \text{ } &L_{val}(w^*(\alpha),\alpha)    \label{eq:outer} \\
  \textbf{s.t. } &w^*(\alpha) = \argmin_w L_{train}(w,\alpha), \text{ }\alpha \in S. \label{eq:inner}
 \end{align}
\end{subequations}
Even with the relaxed space of $\alpha$, the remaining bilevel optimization problem cannot be directly optimized with standard techniques due to the gradient of the outer optimization (\ref{eq:outer}) depending on the gradient of the inner optimization (\ref{eq:inner}), so it is not computationally feasible. Instead, DARTS alternates a single step of gradient descent for $\alpha,$ the outer variable, using either first-order or second-order approximation of the gradient with a single step of gradient descent for $w$, the inner variable. This naive schedule of alternating steps may not be optimal, even in practice: if the parameterized operators are not trained enough at the current architecture encoding for their parameters to be a reasonable approximation of an optima, their architecture weights will be discounted unfairly. Architectures with more skip-connections converge faster even though they tend to have a worse performance at convergence \cite{zhou2020theory}, so this naive schedule may unfairly and undesirably favor skip-connections.  


\begin{figure}[tb!]
\includegraphics[width=\linewidth]{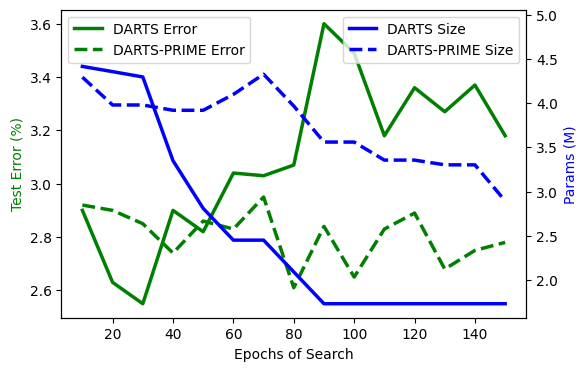}
\caption{Extended trials of DARTS and DARTS-PRIME on CIFAR-10. A single extended trial for each algorithm was run to 150 epochs, and the architecture checkpointed every 10 generations was evaluated once. In DARTS, the architecture eventually has only skip-connections for the searchable part of the architecture, resulting in a high-error performance. DARTS-PRIME avoids this collapse.} \label{fig:long}
\end{figure}

Since the publication of DARTS \cite{liu2018darts}, many variations have been proposed to improve aspects of the original algorithm. We focus on those that are more relevant to the issues we tackle with DARTS-PRIME and study the same DARTS search space in the following.

RobustDARTS \cite{zela2020understanding} investigates failure modes of DARTS, including avoiding the collapse towards excessive skip-connections. Skip-connection mode collapse is a known issue with DARTS, and is demonstrated in Figure \ref{fig:long}. The authors propose several regularizations of the inner optimization (\ref{eq:inner}): in particular, their proposed DARTS-ADA utilizes Hessian information to impose an adaptively increasing regularization, which bares similarities to both our proposed dynamic informed scheduling and increasing proximity regularization. However, this specific variant is not analyzed in the same standard DARTS search space. DARTS+ \cite{liang2019darts+} uses early stopping to avoid skip-connection mode collapse. DARTS- \cite{chu2020darts} avoids skip-connection domination by adding an auxiliary skip-connection in parallel to every edge in addition to the skip-connection operation already being included for each edge in the search. The auxiliary skip-connection is linearly decayed over training, similar to our linear schedule of regularization.

NASP \cite{yao2020efficient} tackles the constraint problem in Differentiable NAS with a modified proximal algorithm \cite{parikh2014proximal}. A partial discretization is applied before each $\alpha$ step and each $w$ step, so forward and backward passes only use and update the current strongest edges of the supernetwork. While this does improve the efficiency of the search without decreasing benchmark performance, it may lead to unselected operators lagging in training compared to selected ones and thus having an unfair disadvantage during search. 


\section{Towards Constrained Optimization}
Neural Architecture Search, particularly when encoded as in DARTS, is a constrained bilevel optimization problem. Both being constrained and being bilevel separately make the optimization less straightforward, such that standard techniques are less applicable. We propose new approaches to the dimensions of bilevel scheduling and regularization towards the constraints to tackle this combined difficult problem in a more dynamic and informed fashion.

\subsection{Dynamic FIMT Scheduling}
Fisher Information of a neural network is closely related to the Hessian matrix of the loss of the network and can be used as a measure of information contained in the weights with respect to the training data as well as a proxy for the steepness of the gradient landscape. While true Fisher Information is expensive to calculate in neural networks, the empirical Fisher Information matrix diagonal can be computed using gradient information already calculated for each batch \cite{martens2016second}. This form of the Fisher Information has been used for updating the momentum term in the Adam optimizer \cite{kingma2014adam}, overcoming catastrophic forgetting \cite{kirkpatrick2017overcoming}, and pruning parameters \cite{tu2016reducing,theis2018faster}.

We compute the trace of the empirical Fisher Information matrix with each minibatch:
\begin{align}
    \bar{F}(w)&=\tr\Big( \nabla_w L_{train}(w,\alpha) \nabla_w L_{train}(w,\alpha)^\mathsf{T} \Big) \label{eq:fimtbig} 
\end{align}
Because the network is changing at each weight update and this measure is subject to variability between minibatches, we use an exponentially weighted moving average of the Fisher Information Matrix Trace (FIMT), similar to the decay factor in the Adam optimizer \cite{kingma2014adam}:
\begin{align}
    \bar{F}_n(w)=\lambda \bar{F}(w) + (1-\lambda) \bar{F}_{n-1}(w), \label{eq:ewmafimt}
\end{align}
for the $n$th minibatch, where $\lambda$ is a hyperparameter.

\begin{figure}[ht!]
\begin{tikzpicture}
  \node (img)  {\includegraphics[width =.95\linewidth]{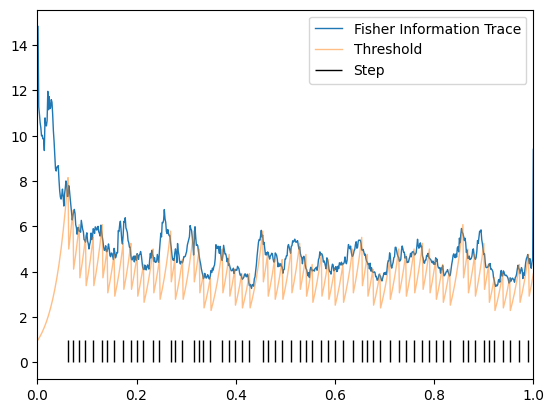}};
  \node[below of=img, node distance=0cm, yshift=-3.0cm,font=\color{black}\scriptsize\sffamily] {Timestep (Epochs)};
  \node[left of=img, node distance=0cm, rotate=90, anchor=center,yshift=3.9cm,font=\color{black}\scriptsize\sffamily] {Value};
 \end{tikzpicture}
\caption{Example of dynamic FIMT schedule, visualized over the first epoch. $\alpha$ updates occur at an interval of roughly every 10 $w$ updates, adjustable by the expected ratio scheduling hyperparameter, but do not occur at the beginning of training when the FIMT is constantly decreasing.}
\label{fig:fimt}
\end{figure}

We use the FIMT as a moving estimate of information contained within the weights of the network and the curvature of the gradient landscape. In particular, as noted in DARTS, when the gradient of $w$ is $0$, then $w$ is a local optimum and thus $w^*=w$, as desired. As a local optima is approached, the FIMT should decrease and $\alpha$ updates can be triggered more frequently. To achieve this effect, we utilize an exponentially adaptive threshold denoted by $h$, parametrized by an initial threshold $h_0$, threshold increasing factor $h_i \geq 1$, expected step ratio $r$, and moving average weighting factor $\lambda.$ A threshold decreasing factor $h_i \leq 1$ is computed such that the expected ratio between $w$ and $\alpha$ updates, given constant $\bar{F}_n(w),$ is equal to $r$. The protocol of this schedule is detailed in \ref{alg:algorithm}. The adaptive threshold causes the $\alpha$ update frequency to increase when the FIMT is generally decreasing over batches and decrease when the FIMT is generally increasing over batches. In turn, this adaptive threshold yields a schedule of $\alpha$ and $w$ updates that is more informed about the convergence of the inner optimization given in Equation \ref{eq:inner} while maintaining efficiency. This avoids unfair discount of relevant architectures. An example of our dynamic FIMT schedule is shown in Figure \ref{fig:fimt}. 

\subsection{Regularization}
In DARTS, the final $\alpha$ values can be far from the resultant discretization, leading to disparity between the search process and its outcome. In order to reward proximity to discretization without losing training signal to inactive portions of the supernetwork, we utilize a proximity regularization that increases in strength over the course of training. The outer objective now becomes
\begin{align}
    &\min_\alpha\Big(L_{val}(w^*(\alpha),\alpha)+\frac{c\rho_{p}}2\norm{(\alpha-\Pi_S(\alpha))}_2\Big),  \label{eq:proxl}
\end{align}
where $c$ increases linearly from $0$ to $1$ over the course of search training , $\rho_{p}$ is a regularization hyperparameter, and $\Pi_S$ is the projection onto set $S$, in this case equivalent to deriving a discretized architecture. This formulation allows for more exploration of $\alpha$ and training of $w$ in parameterized operations particularly at the beginning of the search training, progressively increasing the regularization pressure towards discretization as training continues.

\subsection{DARTS-PRIME}
We combine dynamic FIMT scheduling and proximity regularization with DARTS as DARTS-PRIME. The procedure for DARTS-PRIME is given in Algorithm \ref{alg:algorithm}.

\begin{algorithm}[tb]
\caption{DARTS-PRIME}
\label{alg:algorithm}
 \small
\textbf{Input}: Initial threshold $h_0$, threshold increasing factor $h_i$, expected step ratio $r$, moving average weighting factor $\lambda$, regularization parameter $\rho_p$, and learning rates $\gamma_w$, $\gamma_\alpha$ 

\textbf{Output}: Discrete architecture $\Pi_S(\alpha)$ 

\begin{algorithmic}[1] 
\STATE Initialize $\alpha, w$; let $h\gets h_0, h_d\gets \exp(-r\ln{h_i}), t\gets1$ 
\WHILE{not converged}
\STATE $G \gets \nabla_w L_{train}(w,\alpha)$
\STATE $w \gets w - \gamma_w G$ 
\STATE $\bar{F}_n(w) \gets \lambda \tr\big(G G^\mathsf{T} \big) + (1-\lambda) \bar{F}_{n-1}(w)$
\IF {$\bar{F}_n(w)< h$}
\STATE $\alpha \gets \alpha - \gamma_\alpha \nabla_\alpha \big(L_{val}(w,\alpha) + \frac{c\rho_{p}}2\norm{(\alpha-\Pi_S(\alpha))}_2\big)$
\STATE $h\gets h\cdot h_d$
\ELSE
\STATE $h\gets h\cdot h_i$
\ENDIF
\STATE $n\gets n+1$
\ENDWHILE
\STATE \textbf{return} $\Pi_S(\alpha)$
\end{algorithmic}
\end{algorithm}

In addition to these contributions, we also study a regularization based on the Alternating Direction Method of Multipliers (ADMM), an algorithm that uses dual ascent and the method of multipliers to solve constrained optimization problems \cite{boyd2010}. We propose as well a non-competitive activation function to naturally permit progressive discretization. Further explanation and the results of these studies are presented in Supplementary Section \ref{sec:supppmethod}.

\section{Experiments}
We test the two presented modifications to DARTS together as DARTS-PRIME as well as each independently as ablation studies. Specifically, we test the dynamic $\alpha$ update schedule using Fisher Information (FIMT), and the proximity regularization (PR), in addition to DARTS-PRIME. We compare these methods to DARTS and its variants on three benchmark datasets: the image classification datasets CIFAR-10 and CIFAR-100, and the text prediction dataset Penn TreeBank.

We also rerun DARTS using the code provided by  \cite{liu2018darts}\footnote{https://github.com/quark0/darts}. To better understand the impact of an informed schedule, we study a slight modification of DARTS with a constant schedule of exactly 10 $w$ steps per $\alpha$ step (CS10). The value of 10 was selected for all scheduled experiments as we hypothesize that more $w$ steps should be taken per $\alpha$ step to aid the bilevel optimization approach local minima in $w$ before optimizing $\alpha$.

\begin{table*}[t]
\centering \small
\begin{tabular}{@{}clccc@{}}
\toprule
Task & Method  &  Test Error \hspace{1mm} &  \hspace{1mm} Parameters \hspace{1mm} & \hspace{1mm} Search Time \\
 & &  (\%) \hspace{1mm} &  \hspace{1mm} (M) \hspace{1mm} & \hspace{1mm} (GPU days)
\\  \midrule
\multirow{23}{*}{CIFAR-10} & NASNet-A$^\dagger$ \cite{zoph2017neural} & 2.65 & 3.3 & 1800\\
& AmoebaNet-A*$^\dagger$  \cite{real2019regularized} &  2.55 $\pm$ 0.05  & 2.8 & 3150\\
& PNAS$^\dagger$  \cite{liu2018progressive} & 3.41 $\pm$ 0.09  & 3.2 & 225\\
& ENAS$^\dagger$  \cite{pham2018efficient} & 2.89 & 4.6 & 0.5\\ \cmidrule{2-5} 
& Random Search \cite{liu2018darts} & 3.29 $\pm$ 0.15  & 3.2 & 4\\
& DARTS 1st* \cite{liu2018darts} & 3.00 $\pm$ 0.14  & 3.3 & 1.5\\
& DARTS 2nd* \cite{liu2018darts} & 2.76 $\pm$ 0.09  & 3.3 & 4\\
& P-DARTS \cite{chen2019progressive} & 2.5 & - & 0.3\\
& PD-DARTS \cite{li2020pd} & 2.57 $\pm$ 0.12  & 3.2 & 0.3\\
& FairDARTS \cite{chu2020fair} & 2.54 $\pm$ 0.05 & 3.32 $\pm$ 0.46  & 0.4\\
& R-DARTS (L2) \cite{zela2020understanding} & 2.95 $\pm$ 0.21 & - & 1.6 \\
& DARTS- \cite{chu2020darts} & 2.59 $\pm$ 0.08  & 3.5 $\pm$ 0.13 & 0.4\\
& NASP \cite{yao2020efficient} & 2.83 $\pm$ 0.09 & 3.3 & 0.1\\ \cmidrule{2-5}
& DARTS (our trials) & 2.82 $\pm$ 0.19 & 2.78 $\pm$ 0.28 & 0.4 \\
& +CS10 & 2.80 $\pm$ 0.09 & 3.41 $\pm$ 0.40 & 0.4 \\
& +FIMT & 2.79 $\pm$ 0.12 & 3.49 $\pm$ 0.31 & 0.6 \\
& +PR & 2.77 $\pm$ 0.03 & 3.50 $\pm$ 0.15 & 0.5 \\
& DARTS-PRIME & 2.62 $\pm$ 0.07 & 3.70 $\pm$ 0.27 & 0.5 \\
\midrule\midrule
\multirow{7}{*}{CIFAR-100} & P-DARTS \cite{chen2019progressive} & 15.92  $\pm$ 0.18 & 3.7 & 0.4 \\
& DARTS+ \cite{liang2019darts+} & 15.42 $\pm$ 0.30  & 3.8  & 0.2\\
& R-DARTS (L2) \cite{zela2020understanding} & 18.01 $\pm$ 0.26 & - & -\\
 \cmidrule{2-5}
& DARTS (our trials) & 19.49 $\pm$ 0.60 & 2.02 $\pm$ 0.15 & 0.5 \\
& +FIMT & 19.44 $\pm$ 1.26 & 2.22 $\pm$ 0.22 & 0.5 \\
& +PR & 19.04 $\pm$ 0.85 & 2.24 $\pm$ 0.31 & 0.4 \\
& DARTS-PRIME & 17.44 $\pm$ 0.57 & 3.16 $\pm$ 0.42 & 0.5 \\
\bottomrule
\end{tabular}
\caption{Results for CNN architecture search on CIFAR-10 and CIFAR-100. Test errors (lower is better) are listed in percent of incorrect classifications on the test set after retraining each search architecture. Network sizes of the evaluation network are listed in millions of parameters. Search costs are listed in unnormalized GPU days and do not include evaluation costs. Each of our trials were run on a single V100. Mean errors and search costs are listed with standard deviations. \\ 
*: Previously reported results where the single best search architecture was evaluated multiple times rather than reporting results across multiple searches. Selecting the best architecture across searches is reflected in the search time. \\
$^\dagger$: Architectures searched in a slightly different search space in operator types and network meta-structure.
}
\label{table:res}
\end{table*}

\begin{table*}[ht]
\centering \small
\begin{tabular}{lccccc}
\toprule
\multirow{2}{*}{Method}  & \multicolumn{2}{c}{Perplexity} & Evaluation & Parameters  &  Search Time  \\
 &  Valid  &  Test  & Epochs & (M)  &  (GPU Days) \\  \midrule
NAS$^\dagger$ \cite{zoph2017neural} & - & 64 & - & 25 & 10000\\
ENAS$^{\dagger\ddagger}$ \cite{pham2018efficient} & 60.8 & 58.6 & 8000 & 24 & 0.5 
\\  \midrule
Random search \cite{liu2018darts}  & 61.8 & 59.4 & 8000 & 23 & 2 \\
DARTS 1st \cite{liu2018darts} & 60.2 & 57.6 & 8000 & 23 & 0.5 \\
DARTS 2nd \cite{liu2018darts} & 58.1 & 55.7 & 8000 & 23 & 1 \\ 
GDAS \cite{dong2019searching} & 59.8 & 57.5 & 2000 & 23 & 0.4 \\  \midrule 
DARTS (our trials) & 68.5 $\pm$ 6.2 & 66.2 $\pm$ 6.1 & 1200 & 23 & 0.1 \\
DARTS-PRIME &63.1 $\pm$ 2.1 & 61.1 $\pm$ 2.2 & 1200 & 23 & 0.1 \\
\bottomrule
\end{tabular}
\caption{Results for RNN architecture search on PTB. Performance is measured by perplexity (lower is better). All previous works report the single best result from multiple runs rather than statistics across multiple trials. For our trials, convergence at 1200 epochs was sufficient to differentiate between different methods. \\
$^\dagger$: Architectures searched in a slightly different search space in operator types and network meta-structure. \\
$^\ddagger$: Reevaluated and reported by \cite{liu2018darts} using the current standard evaluation pipeline.}
\label{table:rnn}
\end{table*}

\subsection{Datasets and Tasks}

CIFAR-10 and CIFAR-100 \cite{krizhevsky2009cifar} are image datasets with 10 and 100 classes, respectively, and each consisting of 50K training images and 10K testing images for image classification. Applying the operator and cell-based search space from \cite{liu2018darts} to CIFAR-10 has been heavily studied in related NAS works. We additionally perform both search and evaluation on CIFAR-100 using identical protocols and hyperparameters to explore their applicability to similar tasks.

Penn TreeBank \cite{marcus1993building} (PTB) is a text corpus consisting of over 4.5M American English words for the task of word prediction. We
use the standard pre-processed version of the dataset \cite{pham2018efficient, liu2018darts}. The DARTS search space for RNNs for text prediction searches over activation functions of dense layers within an RNN cell. 

\subsection{DARTS-PRIME Parameters}

We use all protocols and hyperparameters from \cite{liu2018darts} for each domain of image classification and text prediction and for both search and evaluation unless otherwise stated.  We use the first order approximation of the gradient of $\alpha$. Hyperparameters for CIFAR-100 are maintained directly from CIFAR-10. 

When the dynamic FIMT schedule is used, the expected ratio of $w$ steps per $\alpha$ step, $r,$ is set to 10, as in CS10. The initial threshold, $h_0,$ is set to 1.0. The threshold increasing factor, $h_i,$ is set to 1.05. The moving average weighting factor, $\lambda,$ is set to 0.2. These values are all set to intuitive values without formal tuning.

For proximity regularization (PR), the regularization parameter, $\rho_{p},$ is set to 0.1 for image classification and $1.0$ for text prediction. This was exponentially searched over 0.001 to 1.0 on CIFAR-10 and PTB, respectively, during early development.

\subsection{Data and Training Configuration}

After search, we follow the same evaluation procedure as DARTS for image classification and text prediction. After the search training completes and the final discretization selects the cell architecture(s), the evaluation network is reinitialized at the full size, trained on the training set, and tested on the heldout test set.

For trials with the original DARTS schedule of alternating $w$ and $\alpha$ steps, the training and validation split was maintained at 50/50 during search, as done in \cite{liu2018darts}. For trials with either the constant or dynamic schedule with more $w$ steps per $\alpha$ step, the dataset was split according to the expected schedule of steps. For the ratio of 10 $w$ steps per $\alpha$ step, this results in a dataset split of $90.91\%$ in the training set and $9.09\%$ in the validation set.

A difference from the experiment protocol in DARTS is that we do not select the best architecture from multiple searches based on validation performance. Instead, we evaluate each architecture fully once and report the mean test error across search trials with different random seeds, conducting each trial 4 times with random seeds listed in the Supplementary Section \ref{sec:suppconfig}, in addition to more configuration details. 4 was chosen as the number of trials due to the prohibitive cost of the evaluation phase which can take up to 3 GPU days per trial on a V100 GPU. While this is a small number of independent trials, we believe that showing the evaluation results from multiple search trials, instead of the best search architecture, is a better measure of search algorithm performance and reliability. 

\section{Results and Discussion}

We conduct experiments for each of the contributions independently as well as combined with each other in various configurations in the standard DARTS search space for CIFAR-10. Our experiments also include the original DARTS algorithm. We also apply DARTS, DARTS-PRIME, and a subset of ablation configurations to CIFAR-100 and PTB. The mean and standard deviation across searches of the resulting test error or perplexity, final architecture size, and search time for each are reported in Table \ref{table:res} for CIFAR-10 and CIFAR-100 and in Table \ref{table:rnn} for PTB, compared to reported results from related works. The $\alpha$ progressions over training and final architectures for the best trial on CIFAR-10 are shown in Figure \ref{fig:darts} for DARTS and Figure \ref{fig:pf} for DARTS-PRIME. The best architectures for all other experiments are shown in Supplementary Figures.

\begin{figure}[hbtp!]
\begin{minipage}{.9\linewidth}
\centering
\subfloat[Normal cell $\alpha$ over search training]{\includegraphics[trim=5 0 0 20, clip,width = 0.85\linewidth]{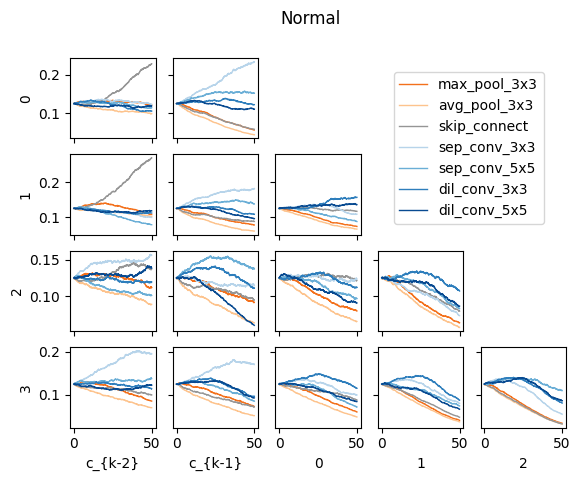}\label{fig:primeNH}}
\end{minipage}\par
\begin{minipage}{.9\linewidth}
\centering
\subfloat[Reduce cell $\alpha$ over search training]{\includegraphics[trim=5 0 0 20, clip,width = 0.85\linewidth]{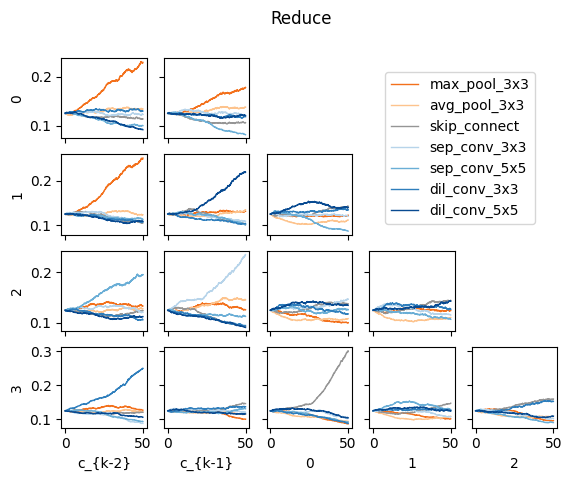}\label{fig:primeRH}}
\end{minipage}\par
\begin{minipage}{.345\linewidth}
\centering
\subfloat[Final normal cell]{\makebox[\linewidth][c]{\includegraphics[width = 0.85\linewidth]{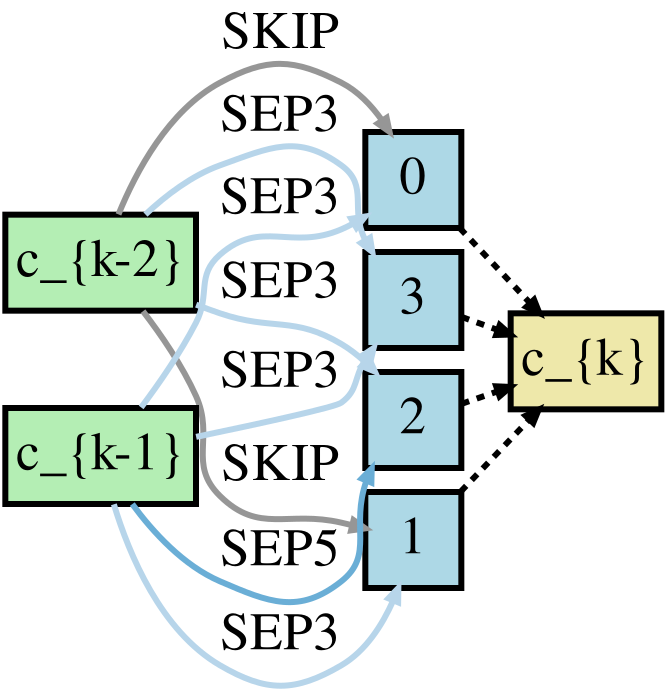}\label{fig:primeNG}}}
\end{minipage}
\begin{minipage}{.645\linewidth}
\centering
\subfloat[Final reduce cell]{\makebox[\linewidth][c]{\includegraphics[width = 0.8\linewidth]{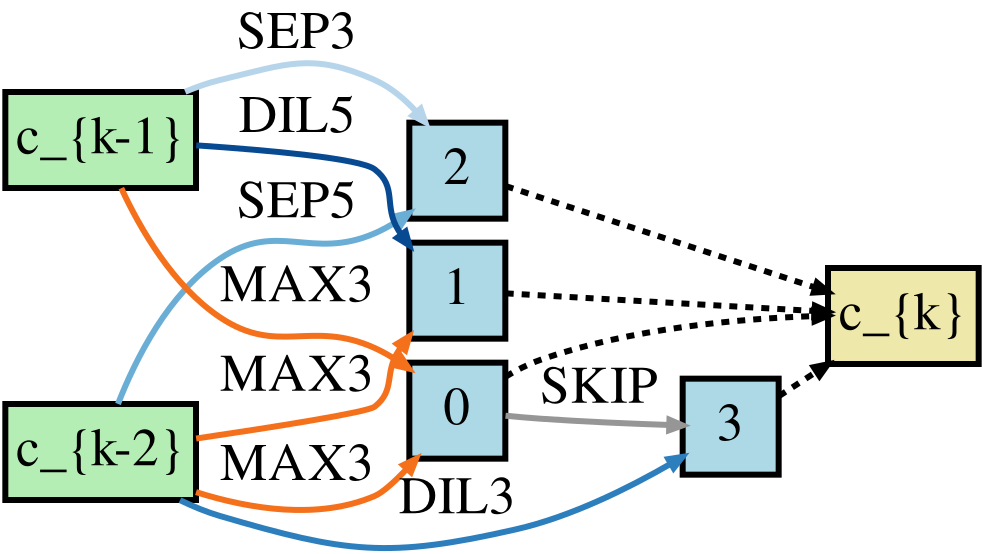}\label{fig:primeRG}}}
\end{minipage}
\caption{Best trial of DARTS-PRIME on CIFAR-10.}
\label{fig:pf}
\end{figure}

The DARTS-PRIME trials show significant improvements over DARTS and match state-of-the-art results within the standard CIFAR-10 search space and task while also yielding one of the lowest variances in performance across our ablation studies. The training progress and final cell architectures for the best trial of DARTS-PRIME are shown in Figure \ref{fig:pf}. This architecture has an evaluation size of 3.76M parameters and a test error of 2.53.

DARTS-PRIME avoids skip-connection mode collapse, as shown in Figure \ref{fig:long}. This is important in domains with variance in parameterization across operators. Avoiding skip-connection mode collapse in the CNN search space decreases the need for tuning the length of search, which improves generalizability across tasks.

We see a clear correlation between network size and performance for this search space in image classification tasks, further visualized in Supplementary Figure \ref{fig:scatter}. This is one of the reasons why the skip-connection mode collapse is catastrophic.

\subsection{Benefits of Scheduling}
Using either CS10 or dynamic FIMT scheduling increases the network size in all paired experiments and tends to improve performance across datasets. This is tied to avoiding the skip-connection mode collapse. Even with many fewer $\alpha$ updates overall, trials using the less frequent schedules have comparable or larger ranges of $\alpha$ values at discretization. Thus, using more informed architecture steps helps prevent unfair suppression of parameterized operations.

Both the constant and dynamic schedules show similar improvement over the original schedule. One characterization of the dynamic schedule is that in the first epoch, the $\alpha$ updates are delayed significantly beyond the expected schedule, as shown in Figure \ref{fig:fimt}. This appears beneficial, as this gives parameterized operations more time to train before $\alpha$ updates start to occur. Characterizing further benefits of dynamic scheduling over constant scheduling is left as future work. Additionally, other information metrics and dynamic scheduling techniques beyond Fisher Information may be explored. 

\subsection{Regularization Improves Reliability}
Proximity regularization yields better and more consistent test error and perplexity results, whether applied alone or with FIMT as in our proposed DARTS-PRIME, especially for CIFAR-10. This is very beneficial for one-shot tasks to improve reliability. 

This implementation of proximity regularization is not the only choice for regularization, particularly towards discretization. In parallel to PR, we developed and conducted a study about another regularization technique for discretization called ADMM regularization. However, our results show that proximity regularization consistently outperforms ADMM regularization. We include these experimental results in the Supplementary Table \ref{table:cares} for the sake of completeness.

The linear scheduling of the increase of proximity regularization strength is a simple implementation choice that yields sufficient performance improvement. It may conflict with applying early stopping as done in other DARTS variants, but only minorly, in a similar way as standard training protocols such as the cosine annealing of architecture weights used in DARTS. This leaves other informed and dynamic schedules, such as similar to our FIMT schedule, as an open question.

\subsection{DARTS-PRIME Performance Beyond CIFAR-10} 

Our findings in CIFAR-10 are also generally supported by the results on CIFAR-100 and PTB: DARTS-PRIME consistently and significantly outperforms DARTS. Specifically on PTB, DARTS-PRIME also has higher reliability in performance. The architecture from our best trial of DARTS-PRIME on CIFAR-100, shown in Supplementary Figure \ref{fig:prime100plots}, has an evaluation size of 3.56M parameters and a test error of 16.82. The architecture from our best trial of DARTS-PRIME on PTB, shown in Supplementary Figure \ref{fig:primernn}, has an validation perplexity of 60.4 and a test perplexity of 58.2 in the shortened evaluation time compared to previous works. 

We note that no further hyperparameter tuning was completed for CIFAR-100, but we anticipate that tuning directly on CIFAR-100 may have allowed DARTS-PRIME to get similar low-variance performance results as the other tasks. Only minor tuning was completed for PTB, which differs from CIFAR in application domain and search space, as the operators of the RNN search space are activation functions, which are not parameterized and do not affect the overall size of the network. The performance of DARTS-PRIME on CIFAR-100 and PTB shows the applicability of the methods and hyperparameters within DARTS-PRIME without extensive searching on target datasets and tasks. 

\section{Conclusion}

In this work, we propose DARTS-PRIME with alternative approaches to two dimensions of differentiable NAS algorithms: informed dynamic scheduling of gradient descent steps and regularization towards discretization. Each of these are analyzed with ablation studies to isolate the effects of each approach across domains and tasks. For practitioners, we recommend dynamic FIMT scheduling for improved network performance and proximity regulation for both improved and more reliable network performance. 

The benefits of these approaches are apparent in results on the CIFAR-10, CIFAR-100, and PTB benchmarks yet also offer insight for different applications of differentiable NAS. The majority of DARTS variants, including this work, focus on narrow and static benchmark problems, achieving similar values of minimal error or perplexity through extensive weight training and hyperparameter tuning. However, more dynamic tasks, such as shifting data distributions or semi-supervised learning as in reinforcement learning, could highlight the benefits of using dynamic optimization schemes, such as our proposed improvements of scheduling and regularization.   

The two improvements of DARTS-PRIME, proximity regularization and a FIMT informed schedule, are modular and could be added to other differentiable NAS algorithms beyond DARTS \cite{liu2018darts}. In particular, proximity regularization may be applied to any algorithm utilizing a continuous relaxation of the architecture, and dynamic FIMT scheduling may be applied to any NAS method posed as a multilevel optimization. While DARTS-PRIME shows a significant improvement over DARTS, we believe that the contributions of a dynamic FIMT scheduling and proximity regularization can be beneficial for one-shot NAS methods in general.

\bibliography{export}

\clearpage

\title{Supplementary for "DARTS-PRIME: Regularization and Scheduling Improve Constrained Optimization in Differentiable NAS"}

\author{Anonymous}

\maketitle
\setcounter{figure}{0}
\renewcommand{\thefigure}{S\arabic{figure}}%
\setcounter{equation}{0}
\renewcommand{\theequation}{S\arabic{equation}}%
\setcounter{table}{0}
\renewcommand{\thetable}{S\arabic{table}}%
\setcounter{section}{0}
\renewcommand{\thesection}{S\arabic{section}}%

\section{Supplementary Methods} \label{sec:supppmethod}

\subsection{CRB Activation}
We propose a new activation of $\alpha$ that improves fairness towards discretization and allows for progressive discretization, called Clipped ReLU Batchnorm (CRB) activation. For CRB activation of $\alpha$, the softmax activation is replaced by unit-clipped ReLU. 

This activation function does not enforce all operator weights within an edge to sum to $1$. So, the distribution of the output of an edge is not as regulated. To account for this in activation distributions, a non-affine batchnorm layer is placed \textit{after} summing across edges for each state within the cell. This permits the $\alpha$ values across edges to still be comparable, as necessary for discretization. Additionally, the "none" operator can be removed, since the activated $\alpha$ values are no longer dependent on each other.

CRB activation provides bounds within $[0,1]$ and a natural progressive pruning heuristic. Over training, when any architecture weight becomes non-positive, the corresponding operator is pruned from that edge. This progressive pruning allows the network training to focus on the remaining operators and saves computational time.

\subsection{ADMM Regularization}
The Alternating Direction Method of Multipliers (ADMM) is an algorithm that uses dual ascent and the method of multipliers to solve constrained optimization problems \cite{boyd2010}. ADMM has previously been applied to training sparse networks \cite{kiaee2016alternating}, pruning weights \cite{zhang2018structadmm,ye2018progressive}, and pruning channels \cite{ma2019}, using the version of the algorithm designed for nonconvex constraints such as maximal cardinality of weight matrices. When applied in practice the non-convex objective of neural network objective functions, the optimization variable of the network parameters is usually not evaluated to convergence within every iteration. The implementation of ADMM functions similarly to a dynamic regularization, so we apply it to differentiable NAS in comparison to our proposed proximity regularization.

Directly applying ADMM for nonconvex constraints \cite{boyd2010} to the optimization given in Problem \ref{eq:bilevel} using $\alpha \in S$ as the constraint, the iterative solution is
\begin{align}
    \alpha^{k+1} := & \argmin_\alpha\Big(L_{val}(w^*(\alpha),\alpha)+\frac{\rho_{a}}2\norm{\alpha-z^k+u^k}_2^2\Big) \label{eq:admma}\\
    &\textbf{ s.t. } w^*(\alpha) = \argmin_w L_{train}(w,\alpha) \label{eq:admmw}\\
    z^{k+1} := &\Pi_S(\alpha^{k+1}+u^k) \\
    u^{k+1} := &u^k+\alpha^{k+1} - z^{k+1}, \label{eq:u}
\end{align}
where $\Pi_S$ is the projection onto $S$, $z$ and $u$ are introduced variables, and $\rho_{a}$ is a regularization hyperparameter.

Equation \ref{eq:admma} is solved by using the DARTS $1^\text{st}$ order approximation for $n$ minibatches before updating $z$ and $u$. The number of $w$ steps (Equation \ref{eq:admmw}) for each $\alpha$ step (Equation \ref{eq:admma}) is determined by the $\alpha$ schedule.

Because neither level of the bilevel optimization is evaluated to convergence within each iteration, we discount past $u$ values with each update using a discount factor, $\lambda_u$. Thus, Equation \ref{eq:u} becomes
\begin{align}
    u^{k+1} := &\lambda_u u^k+\alpha^{k+1} - z^{k+1}.
\end{align}

\section{Supplementary Configurations} \label{sec:suppconfig}

All experiments were run using the code here: https://anonymous.4open.science/r/DARTS-PRIME. All Python packages used in this repository are open-source, and current versions of each were used. Each trial was run with a single V100 GPU on a computing node running Linux. The random seeds used were 101, 102, 103, and 104.

\section{Supplementary Results} \label{sec:suppresults}
In our CIFAR-10 experiments for ADMM regularization, $z$ and $u$ are updated after every 10 $\alpha$ steps. The decay factor, $\lambda_u,$ is set to $0.8$. The regularization parameter, $\rho_{a},$ is set to 0.1. When CRB activation is used, the cosine annealing of the architecture learning rate and the weight decay of $\alpha$ are both turned off.

\begin{table*}[tb!]
\centering \small
\begin{tabular}{@{}clccc@{}}
\toprule
Task & Method  &  Test Error \hspace{1mm} &  \hspace{1mm} Params \hspace{1mm} & \hspace{1mm} Search Time \\
 & &  (\%) \hspace{1mm} &  \hspace{1mm} (M) \hspace{1mm} & \hspace{1mm} (GPU days)
\\  \midrule
\multirow{6}{*}{CIFAR-10} 
& +CRB & 3.58 $\pm$ 0.79 & 1.77 $\pm$ 0.25 & 0.4 \\
& +FIMT+CRB & 2.91 $\pm$ 0.13 & 2.34 $\pm$ 0.35 & 0.5 \\
& DARTS-PRIME +CRB & 2.98 $\pm$ 0.22 & 3.20 $\pm$ 0.33 & 0.5 \\
& +ADMM & 2.89 $\pm$ 0.05 & 3.13 $\pm$ 0.30 & 0.4 \\
& +ADMM+FIMT & 2.85 $\pm$ 0.15 & 4.43 $\pm$ 0.99 & 0.5 \\
& +ADMM+FIMT+CRB & 3.18 $\pm$ 0.29 & 2.06 $\pm$ 0.26 & 0.4 \\
\midrule\midrule
\multirow{1}{*}{CIFAR-100} & DARTS-PRIME +CRB & 18.41 $\pm$ 0.48 & 2.28 $\pm$ 0.24 & 0.4 \\
\bottomrule
\end{tabular}
\caption{Results for our variants using ADMM regularization on CIFAR-10.
}
\label{table:cares}
\end{table*}

\subsection{CRB Activation Produces Smaller Networks}

\begin{figure}[b!]
\centering
\begin{minipage}{\linewidth}
\centering
\subfloat[]{\includegraphics[trim=30 5 40 40, clip, width=\linewidth]{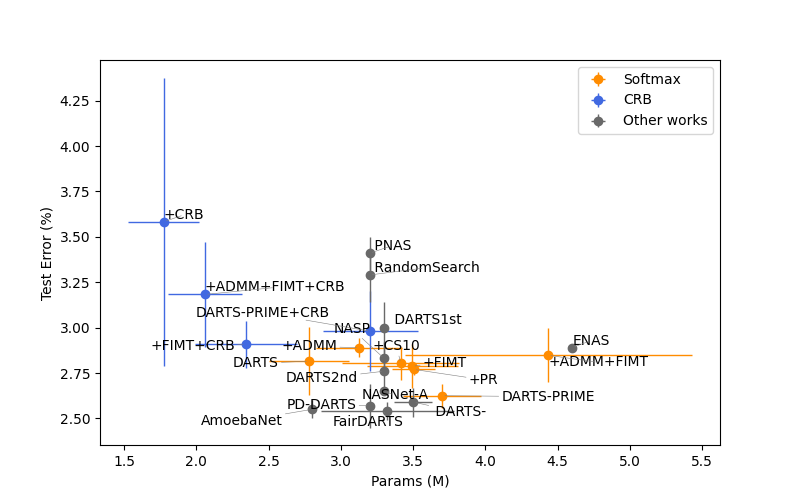}\label{fig:scatAct}} 
\end{minipage}\par
\begin{minipage}{\linewidth}
\centering
\subfloat[]{\includegraphics[trim=30 5 40 40, clip, width=\linewidth]{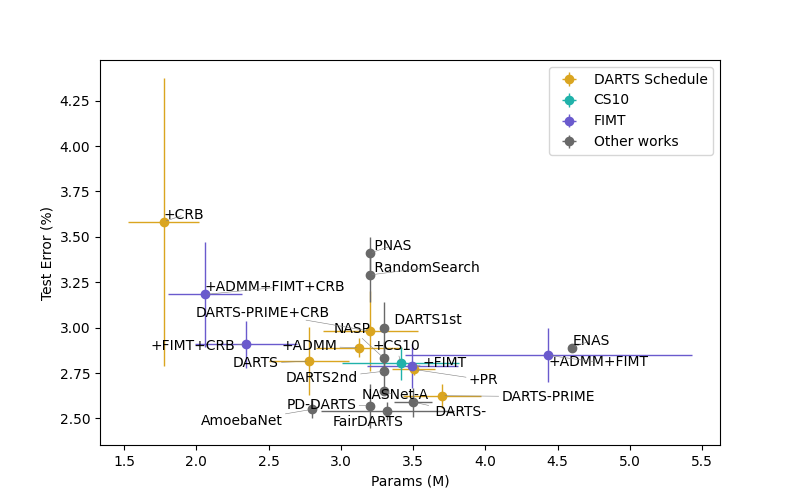}\label{fig:scatSched}} 
\end{minipage}\par
\begin{minipage}{\linewidth}
\centering
\subfloat[]{\includegraphics[trim=30 5 40 40, clip, width=\linewidth]{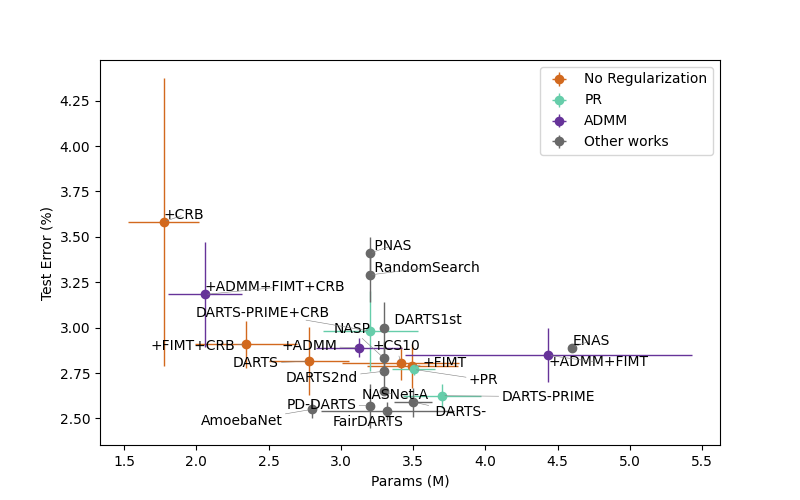}\label{fig:scatReg}} 
\end{minipage}\par
\caption{Comparison across CIFAR-10 trials of test error vs. network size, recolored by (a) activation, (b) schedule, and (c) regularization. } \label{fig:scatter}
\end{figure}

For all combinations, the CRB activation finds smaller networks than the corresponding experiment with softmax activation, albeit with a small degradation in performance. This pattern is shown in Figure \ref{fig:scatAct}. We noted that only trials using CRB activation selected any pooling operators in the normal cell architecture, where other methods use convolutional operators. This could cause performance degradation due to the difference in depth between the search and evaluation networks, particularly since only normal cells are added to reach the final depth. 

We also note that CRB activation has high variance without regulation or informed scheduling. Particularly in the DARTS+CRB experiment, one trial had a very high error of 4.95, which greatly increased the mean and variance. This trial fell into skip-connection mode collapse. As discussed in the following, scheduling and especially regularization help prevent this collapse from occurring, which is beneficial if the search is applied in a truly one-shot task. On the other hand, if multiple searches can be completed and small network size is beneficial to the application, this protocol may actually be the best approach to apply.

\begin{figure}[thp!]
\begin{minipage}{\linewidth}
\centering
\subfloat[Normal cell $\alpha$ over search training]{\includegraphics[trim=5 0 0 20, clip,width = 2.9in]{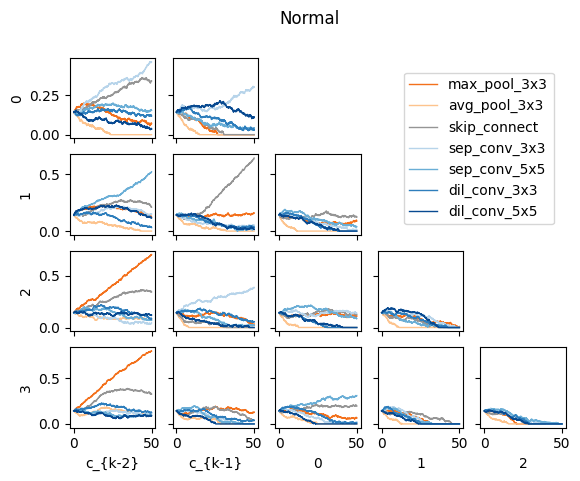}\label{fig:prime+cNH}}
\end{minipage}\par
\begin{minipage}{\linewidth}
\centering
\subfloat[Reduce cell $\alpha$ over search training]{\includegraphics[trim=5 0 0 20, clip,width = 2.9in]{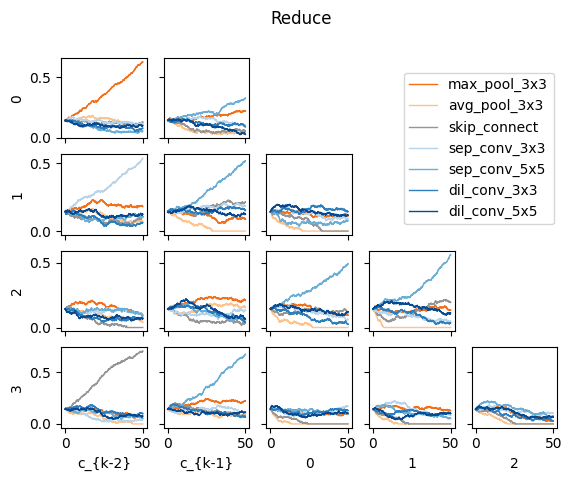}\label{fig:prime+cRH}}
\end{minipage}\par
\begin{minipage}{.49\linewidth}
\centering
\subfloat[Final normal cell]{\makebox[\linewidth][c]{\includegraphics[width = \linewidth]{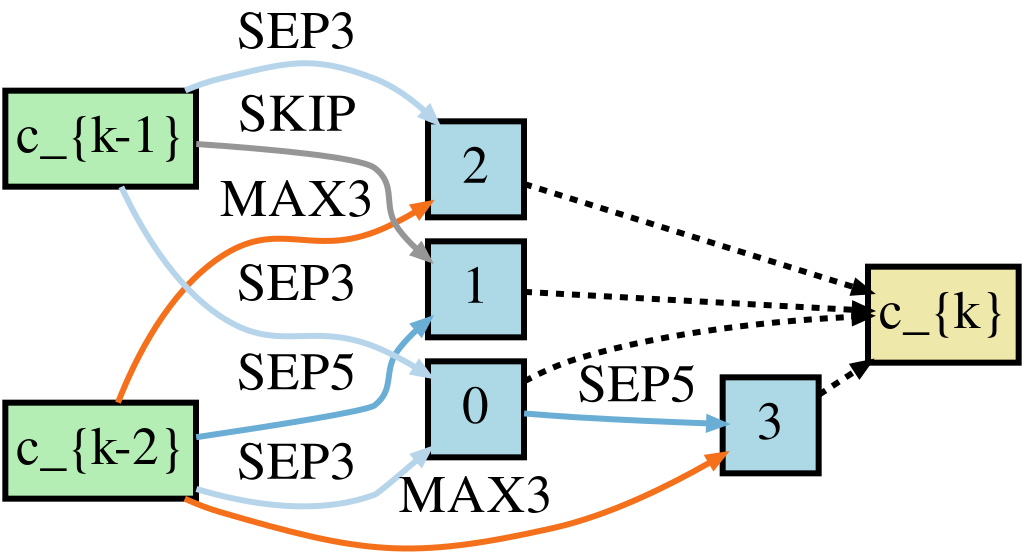}\label{fig:prime+cHG}}}
\end{minipage}
\begin{minipage}{.49\linewidth}
\centering
\subfloat[Final reduce cell]{\makebox[\linewidth][c]{\includegraphics[width = \linewidth]{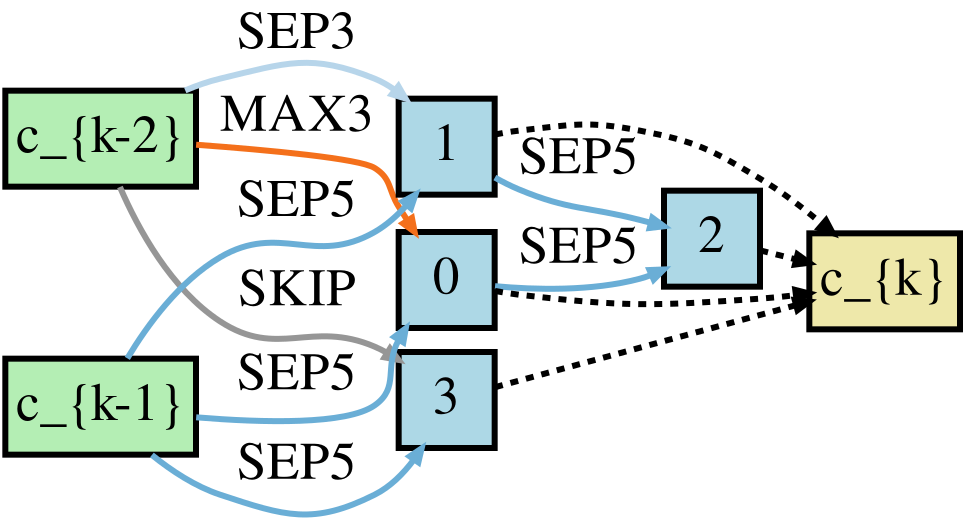}\label{fig:prime+cRG}}}
\end{minipage}
\caption{Best trial of DARTS-PRIME +CRB.}
\label{fig:pfc}
\end{figure}

An example of progressive pruning provided by CRB activation is shown in Figure \ref{fig:pfc}(a-b). 

\subsection{ADMM Regularization Improves Performance}

Using ADMM regularization generally improves performance in each paired experiment. However, proximity regularization was more performant in all cases. ADMM is a very powerful algorithm but with many hyperparameters that require careful tuning when applied in practice. This leaves further development as open work.

\section{Supplementary Figures}
\begin{figure}[thp!]
\centering
\begin{minipage}{.39\linewidth}
\centering
\subfloat[Final normal cell]{\makebox[\linewidth][c]{\includegraphics[width = \linewidth]{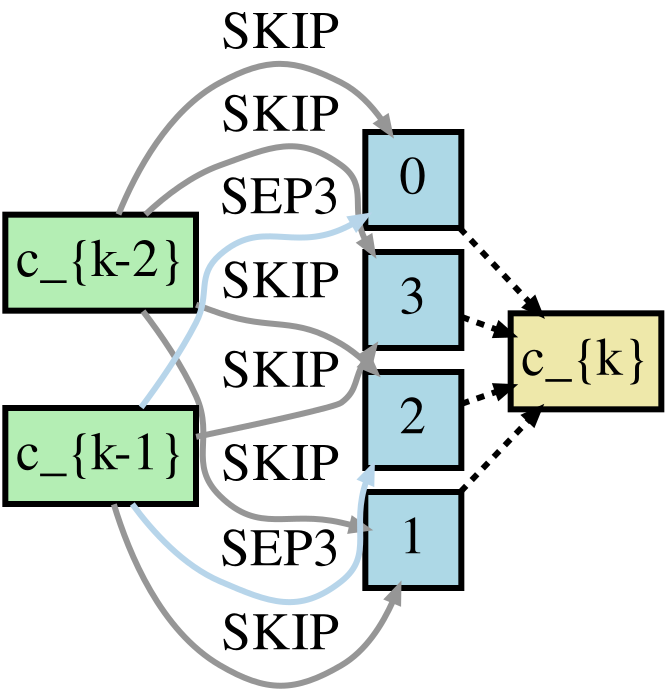}\label{fig:darts100NG}}}
\end{minipage}
\begin{minipage}{.59\linewidth}
\centering
\subfloat[Final reduce cell]{\makebox[\linewidth][c]{\includegraphics[width = \linewidth]{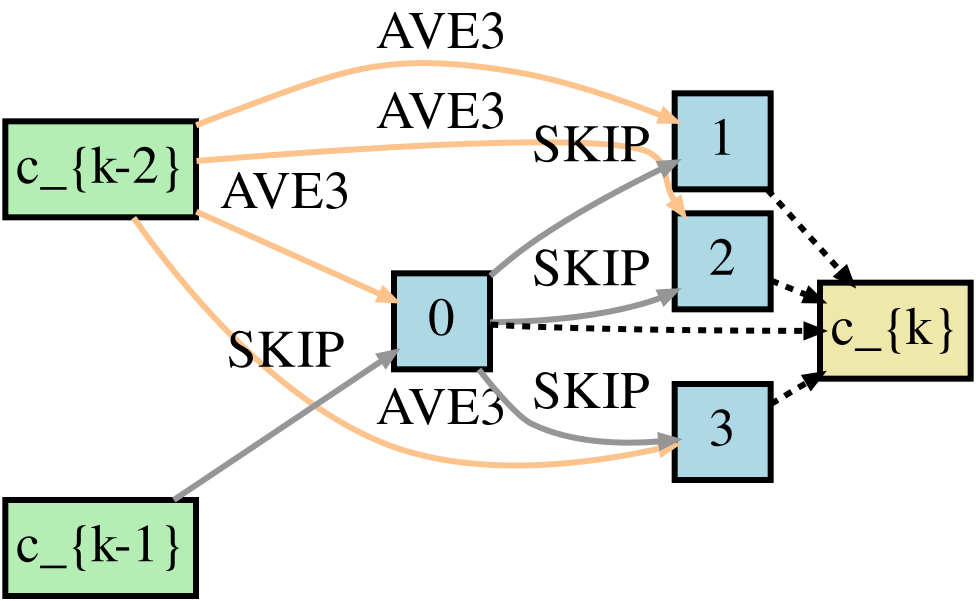}\label{fig:darts100RG}}}
\end{minipage}
\caption{Best final architecture of DARTS on CIFAR-100.}
\label{fig:darts100plots}
\end{figure}

\begin{figure}[thp!]
\centering
\begin{minipage}{.49\linewidth}
\centering
\subfloat[Final normal cell]{\makebox[\linewidth][c]{\includegraphics[width = \linewidth]{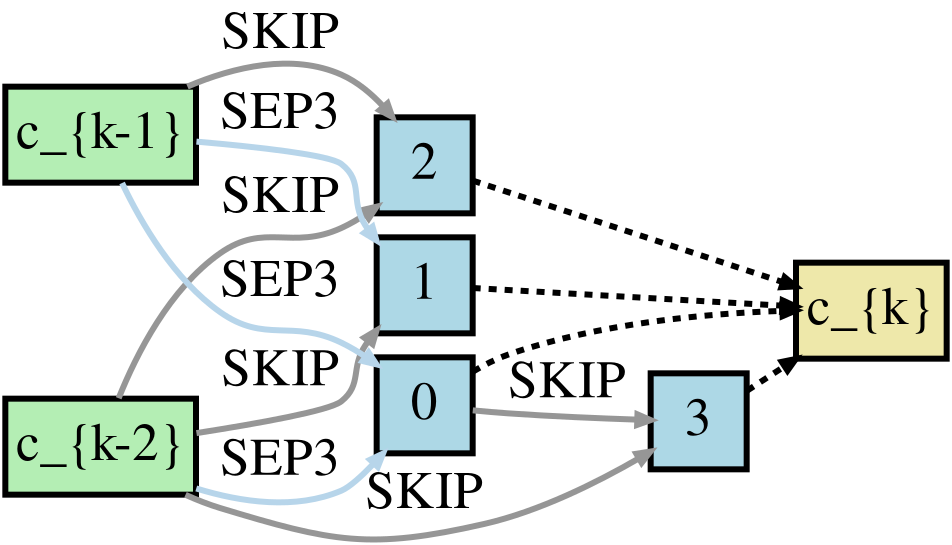}\label{fig:prime100NG}}}
\end{minipage}
\begin{minipage}{.49\linewidth}
\centering
\subfloat[Final reduce cell]{\makebox[\linewidth][c]{\includegraphics[width = \linewidth]{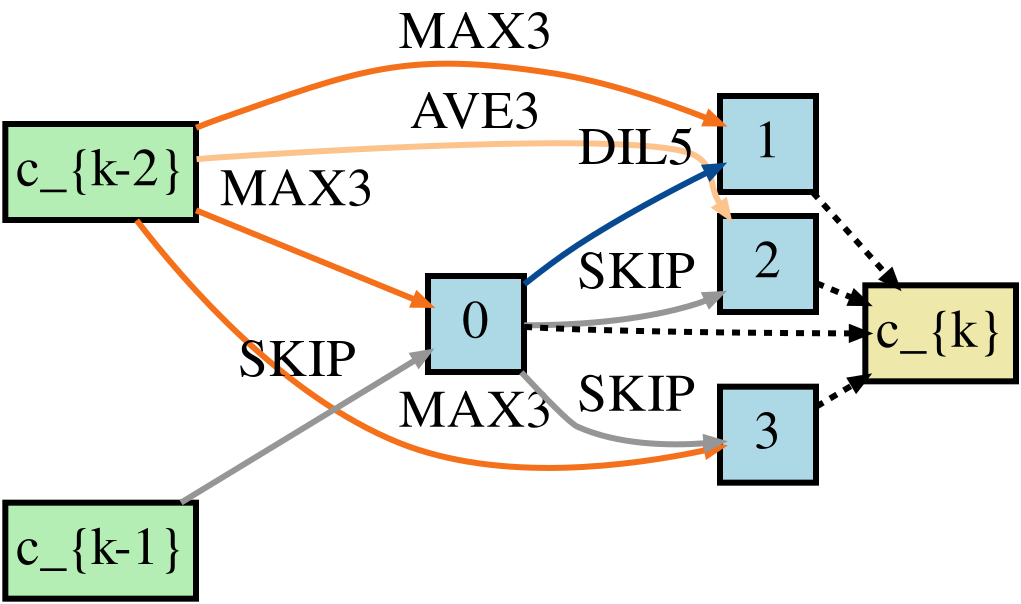}\label{fig:prime100RG}}}
\end{minipage}
\caption{Best final architecture of DARTS-PRIME on CIFAR-100.}
\label{fig:prime100plots}
\end{figure}

\begin{figure}[thp!]
\includegraphics[width=\linewidth]{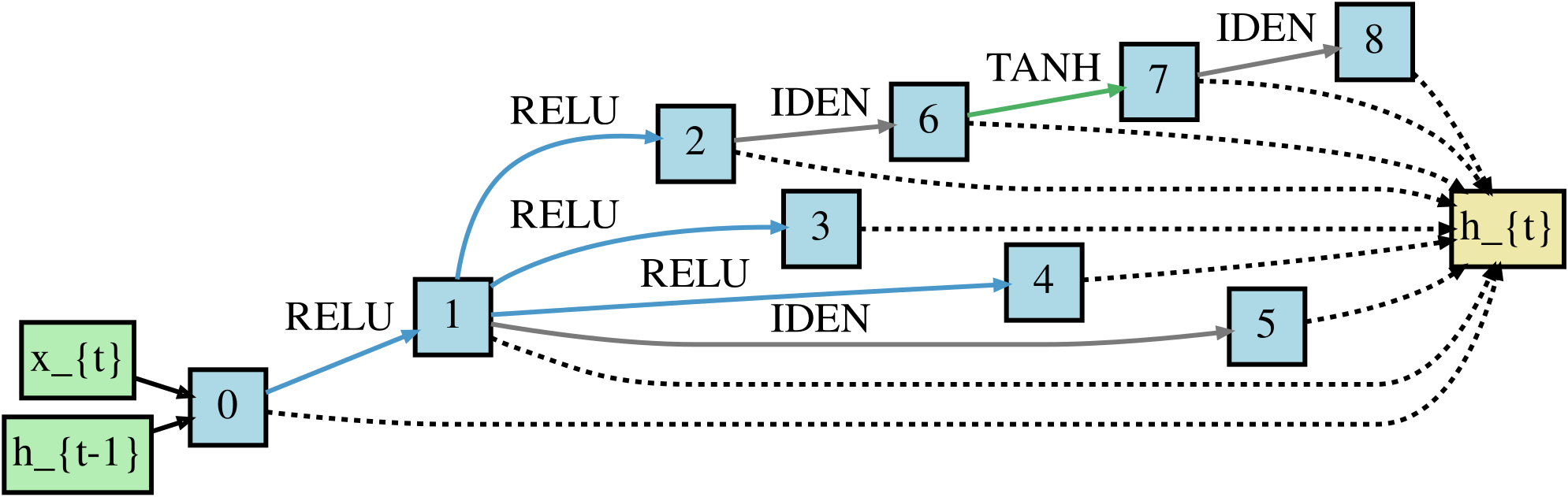}
\caption{Best final architecture of DARTS on PTB.} \label{fig:dartsrnn}
\end{figure}

\begin{figure}[thp!]
\includegraphics[width=\linewidth]{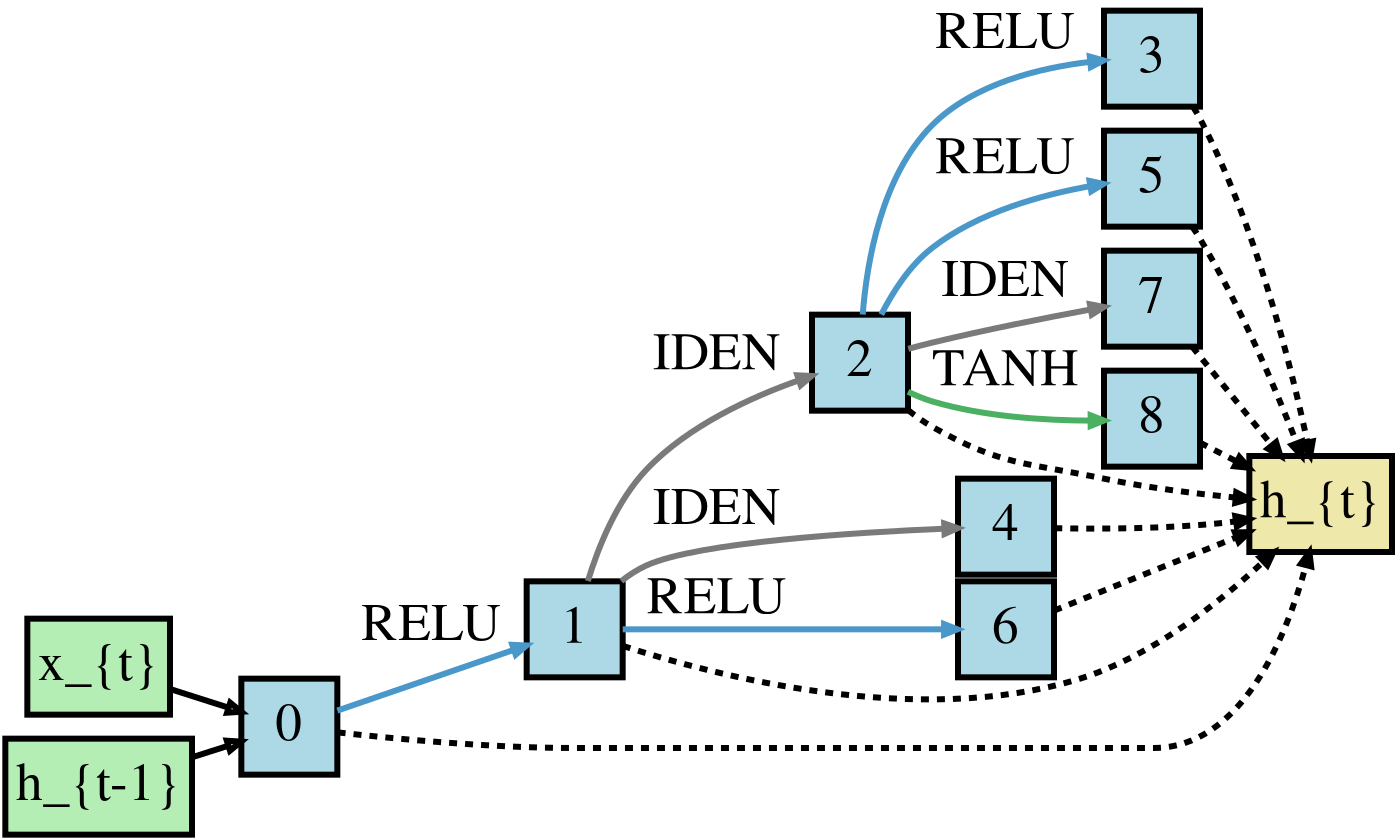}
\caption{Best final architecture of DARTS-PRIME on PTB.} \label{fig:primernn}
\end{figure}

\end{document}


\maketitle
\setcounter{figure}{0}
\renewcommand{\thefigure}{S\arabic{figure}}%
\setcounter{equation}{0}
\renewcommand{\theequation}{S\arabic{equation}}%

\section{Supplementary methods}

\subsection{CRB Activation of $\alpha$}
We propose a new activation of $\alpha$ that improves fairness towards discretization and allows for progressive discretization, called Clipped ReLU Batchnorm (CRB) activation. For CRB activation of $\alpha$, the softmax activation is replaced by unit-clipped ReLU. 

This activation function does not enforce all operator weights within an edge to sum to $1$. So, the distribution of the output of an edge is not as regulated. To account for this in activation distributions, a non-affine batchnorm layer is placed \textit{after} summing across edges for each state within the cell. This permits the $\alpha$ values across edges to still be comparable, as necessary for discretization. Additionally, the "none" operator can be removed, since the activated $\alpha$ values are no longer dependent on each other.

CRB activation provides bounds within $[0,1]$ and a natural progressive pruning heuristic. Over training, when any architecture weight becomes non-positive, the corresponding operator is pruned from that edge. This progressive pruning allows the network training to focus on the remaining operators and saves computational time.

\subsection{ADMM Regularization}
The Alternating Direction Method of Multipliers (ADMM) is an algorithm that uses dual ascent and the method of multipliers to solve constrained optimization problems \cite{boyd2010}. ADMM has previously been applied to training sparse networks \cite{kiaee2016alternating}, pruning weights \cite{zhang2018structadmm,ye2018progressive}, and pruning channels \cite{ma2019}, using the version of the algorithm designed for nonconvex constraints such as maximal cardinality of weight matrices. When applied in practice the non-convex objective of neural network objective functions, the optimization variable of the network parameters is usually not evaluated to convergence within every iteration. The implementation of ADMM functions similarly to a dynamic regularization, so we apply it to differentiable NAS in comparison to our proposed proximity regularization.

Directly applying ADMM for nonconvex constraints \cite{boyd2010} to the optimization given in Problem \ref{main-eq:bilevel} using $\alpha \in S$ as the constraint, the iterative solution is
\begin{align}
    \alpha^{k+1} := & \argmin_\alpha\Big(L_{val}(w^*(\alpha),\alpha)+\frac{\rho_{a}}2\norm{\alpha-z^k+u^k}_2^2\Big) \label{eq:admma}\\
    &\textbf{ s.t. } w^*(\alpha) = \argmin_w L_{train}(w,\alpha) \label{eq:admmw}\\
    z^{k+1} := &\Pi_S(\alpha^{k+1}+u^k) \\
    u^{k+1} := &u^k+\alpha^{k+1} - z^{k+1}, \label{eq:u}
\end{align}
where $\Pi_S$ is the projection onto $S$, $z$ and $u$ are introduced variables, and $\rho_{a}$ is a regularization hyperparameter.

Equation \ref{eq:admma} is solved by using the DARTS $1^\text{st}$ order approximation for $n$ minibatches before updating $z$ and $u$. The number of $w$ steps (Equation \ref{eq:admmw}) for each $\alpha$ step (Equation \ref{eq:admma}) is determined by the $\alpha$ schedule.

Because neither level of the bilevel optimization is evaluated to convergence within each iteration, we discount past $u$ values with each update using a discount factor, $\lambda_u$. Thus, Equation \ref{eq:u} becomes
\begin{align}
    u^{k+1} := &\lambda_u u^k+\alpha^{k+1} - z^{k+1}.
\end{align}

\section{Supplementary Results}
In our CIFAR-10 experiments for ADMM regularization, $z$ and $u$ are updated after every 10 $\alpha$ steps. The decay factor, $\lambda_u,$ is set to $0.8$. The regularization parameter, $\rho_{a},$ is set to 0.1. When CRB activation is used, the cosine annealing of the architecture learning rate and the weight decay of $\alpha$ are both turned off.

\begin{table*}[t]
\centering \small
\begin{tabular}{@{}clccc@{}}
\toprule
Task & Method  &  Test Error \hspace{1mm} &  \hspace{1mm} Params \hspace{1mm} & \hspace{1mm} Search Time \\
 & &  (\%) \hspace{1mm} &  \hspace{1mm} (M) \hspace{1mm} & \hspace{1mm} (GPU days)
\\  \midrule
\multirow{6}{*}{CIFAR-10} 
& +CRB & 3.58 $\pm$ 0.79 & 1.77 $\pm$ 0.25 & 0.4 \\
& +FIMT+CRB & 2.91 $\pm$ 0.13 & 2.34 $\pm$ 0.35 & 0.5 \\
& DARTS-PRIME +CRB & 2.98 $\pm$ 0.22 & 3.20 $\pm$ 0.33 & 0.5 \\
& +ADMM & 2.89 $\pm$ 0.05 & 3.13 $\pm$ 0.30 & 0.4 \\
& +ADMM+FIMT & 2.85 $\pm$ 0.15 & 4.43 $\pm$ 0.99 & 0.5 \\
& +ADMM+FIMT+CRB & 3.18 $\pm$ 0.29 & 2.06 $\pm$ 0.26 & 0.4 \\
\midrule\midrule
\multirow{1}{*}{CIFAR-100} & DARTS-PRIME+CRB & 18.41 $\pm$ 0.48 & 2.28 $\pm$ 0.24 & 0.4 \\
\bottomrule
\end{tabular}
\caption{Results for our variants using ADMM regularization on CIFAR-10.
}
\label{table:res}
\end{table*}

\subsection{CRB Activation Produces Smaller Networks}

For all combinations, the CRB activation finds smaller networks than the corresponding experiment with softmax activation, albeit with a small degradation in performance. This pattern is shown in Figure \ref{fig:scatCRB}. We noted that only trials using CRB activation selected any pooling operators in the normal cell architecture, where other methods use convolutional operators. This could cause performance degradation due to the difference in depth between the search and evaluation networks, particularly since only normal cells are added to reach the final depth. 

We also note that CRB activation has high variance without regulation or informed scheduling. Particularly in the DARTS+CRB experiment, one trial had a very high error of 4.95, which greatly increased the mean and variance. This trial fell into skip connection mode collapse. As discussed in the following, scheduling and especially regularization help prevent this collapse from occurring, which is beneficial if the search is applied in a truly one-shot task. On the other hand, if multiple searches can be completed and small network size is beneficial to the application, this protocol may actually be the best approach to apply.

An example of progressive pruning provided by CRB activation is shown in Figure \ref{fig:pfc}(a-b). 

\section{Supplementary figures}
\begin{figure}[b!]
\centering
\begin{minipage}{\linewidth}
\centering
\subfloat[]{\includegraphics[trim=30 5 40 40, clip, width=\linewidth]{images/scatterCRB.png}\label{fig:scatSched}} 
\end{minipage}\par
\begin{minipage}{\linewidth}
\centering
\subfloat[]{\includegraphics[trim=30 5 40 40, clip, width=\linewidth]{images/scatterFIMT.png}\label{fig:scatSched}} 
\end{minipage}\par
\begin{minipage}{\linewidth}
\centering
\subfloat[]{\includegraphics[trim=30 5 40 40, clip, width=\linewidth]{images/scatterADMM.png}\label{fig:scatReg}} 
\end{minipage}\par
\caption{Comparison across trials of test error vs network size, recolored by (a) activation, (b) schedule, and (c) regularization. } \label{fig:scatter}
\end{figure}

\begin{figure*}
\begin{tabular}{cc}
\centering
\subfloat[Normal cell $\alpha$ over search training]{\includegraphics[trim=5 0 0 20, clip, width = .5\linewidth]{images/darts10/Normalalpha_alphahistory-none.png}\label{fig:darts10NH}} &
\subfloat[Reduce cell $\alpha$ over search training]{\includegraphics[trim=5 0 0 20, clip, width = .5\linewidth]{images/darts10/Reducealpha_alphahistory-none.png}\label{fig:darts10RH}}
\\
\subfloat[Final normal cell architecture]{\makebox[.5\linewidth][c]{\includegraphics[height = 0.9in]{images/darts10/normalgraphup.png}\label{fig:darts10NG}}}
 &
\subfloat[Final reduce cell architecture]{\makebox[.5\linewidth][c]{\includegraphics[height = 0.9in]{images/darts10/reducegraphup.png}\label{fig:darts10RG}}}
\\
\subfloat[Normal cell $\alpha$ over search training]{\includegraphics[trim=5 0 0 20, clip, width = .5\linewidth]{images/dartsfimt/Normalalpha_alphahistory-none.png}\label{fig:dartsfimtNH}} 
& 
\subfloat[Reduce cell $\alpha$ over search training]{\includegraphics[trim=5 0 0 20, clip, width = .5\linewidth]{images/dartsfimt/Reducealpha_alphahistory-none.png}\label{fig:dartsfimtRH}} 
\\
\subfloat[Final normal cell architecture]{\makebox[.5\linewidth][c]{\includegraphics[height = 0.9in]{images/dartsfimt/normalgraphup.png}\label{fig:dartsfimtHG}}}
&
\subfloat[Final reduce cell architecture]{\makebox[.5\linewidth][c]{\includegraphics[height = 0.9in]{images/dartsfimt/reducegraphup.png}\label{fig:dartfimtRG}}}
\end{tabular}
\caption{Best trials of +CS10 (a-d) and +FIMT (e-h).}
\label{fig:10FIMT}
\end{figure*}

\begin{figure*}
\begin{tabular}{cc}
\centering
\subfloat[Normal cell $\alpha$ over search training]{\includegraphics[trim=5 0 0 20, clip, width = .5\linewidth]{images/dartscrb/Normalalpha_history.png}\label{fig:dartscrbNH}} &
\subfloat[Reduce cell $\alpha$ over search training]{\includegraphics[trim=5 0 0 20, clip, width = .5\linewidth]{images/dartscrb/Reducealpha_history.png}\label{fig:dartscrbRH}}
\\
\subfloat[Final normal cell architecture]{\makebox[.5\linewidth][c]{\includegraphics[height = 0.9in]{images/dartscrb/normalgraphup.png}\label{fig:dartscrbNG}}}
 &
\subfloat[Final reduce cell architecture]{\makebox[.5\linewidth][c]{\includegraphics[height = 0.9in]{images/dartscrb/reducegraphup.png}\label{fig:dartscrbRG}}}
\\
\subfloat[Normal cell $\alpha$ over search training]{\includegraphics[trim=5 0 0 20, clip, width = .5\linewidth]{images/dartsfc/Normalalpha_history.png}\label{fig:dartsfcNH}} 
& 
\subfloat[Reduce cell $\alpha$ over search training]{\includegraphics[trim=5 0 0 20, clip, width = .5\linewidth]{images/dartsfc/Reducealpha_history.png}\label{fig:dartsfcRH}} 
\\
\subfloat[Final normal cell architecture]{\makebox[.5\linewidth][c]{\includegraphics[height = 0.9in]{images/dartsfc/normalgraphup.png}\label{fig:dartsfcHG}}}
&
\subfloat[Final reduce cell architecture]{\makebox[.5\linewidth][c]{\includegraphics[height = 0.9in]{images/dartsfc/reducegraphup.png}\label{fig:dartfcRG}}}
\end{tabular}
\caption{Best trials of +CRB (a-d) and +FIMT+CRB (e-h).}
\label{fig:crbfc}
\end{figure*}

\begin{figure*}
\begin{tabular}{cc}
\centering
\subfloat[Normal cell $\alpha$ over search training]{\includegraphics[trim=5 0 0 20, clip, width = .5\linewidth]{images/proxbase/Normalalpha_alphahistory-none.png}\label{fig:proxbaseNH}} &
\subfloat[Reduce cell $\alpha$ over search training]{\includegraphics[trim=5 0 0 20, clip, width = .5\linewidth]{images/proxbase/Reducealpha_alphahistory-none.png}\label{fig:proxbaseRH}}
\\
\subfloat[Final normal cell architecture]{\makebox[.5\linewidth][c]{\includegraphics[height = 0.9in]{images/proxbase/normalgraphup.png}\label{fig:proxbaseNG}}}
 &
\subfloat[Final reduce cell architecture]{\makebox[.5\linewidth][c]{\includegraphics[height = 0.9in]{images/proxbase/reducegraphup.png}\label{fig:proxbaseRG}}}
\\
\subfloat[Normal cell $\alpha$ over search training]{\includegraphics[trim=5 0 0 20, clip, width = .5\linewidth]{images/admmbase/Normalalpha_alphahistory-none.png}\label{fig:admmbaseNH}} 
& 
\subfloat[Reduce cell $\alpha$ over search training]{\includegraphics[trim=5 0 0 20, clip, width = .5\linewidth]{images/admmbase/Reducealpha_alphahistory-none.png}\label{fig:admmbaseRH}} 
\\
\subfloat[Final normal cell architecture]{\makebox[.5\linewidth][c]{\includegraphics[height = 0.9in]{images/admmbase/normalgraphup.png}\label{fig:admmbaseHG}}}
&
\subfloat[Final reduce cell architecture]{\makebox[.5\linewidth][c]{\includegraphics[height = 0.9in]{images/admmbase/reducegraphup.png}\label{fig:admmbaseRG}}}
\end{tabular}
\caption{Best trials of +PR (a-d) and +ADMM (e-h).}
\label{fig:prox_admmbase}
\end{figure*}

\begin{figure*}
\begin{tabular}{cc}
\centering
\subfloat[Normal cell $\alpha$ over search training]{\includegraphics[trim=5 0 0 20, clip, width = .5\linewidth]{images/admmfimt/Normalalpha_alphahistory-none.png}\label{fig:admmfimtNH}} &
\subfloat[Reduce cell $\alpha$ over search training]{\includegraphics[trim=5 0 0 20, clip, width = .5\linewidth]{images/admmfimt/Reducealpha_alphahistory-none.png}\label{fig:admmfimtRH}}
\\
\subfloat[Final normal cell architecture]{\makebox[.5\linewidth][c]{\includegraphics[height = 0.9in]{images/admmfimt/normalgraphup.png}\label{fig:admmfimtNG}}}
 &
\subfloat[Final reduce cell architecture]{\makebox[.5\linewidth][c]{\includegraphics[height = 0.9in]{images/admmfimt/reducegraphup.png}\label{fig:admmfimtRG}}}
\\
\subfloat[Normal cell $\alpha$ over search training]{\includegraphics[trim=5 0 0 20, clip, width = .5\linewidth]{images/admm/Normalalpha_history.png}\label{fig:admmNH}} 
& 
\subfloat[Reduce cell $\alpha$ over search training]{\includegraphics[trim=5 0 0 20, clip, width = .5\linewidth]{images/admm/Reducealpha_history.png}\label{fig:admmRH}} 
\\
\subfloat[Final normal cell architecture]{\makebox[.5\linewidth][c]{\includegraphics[height = 0.9in]{images/admm/normalgraphup.png}\label{fig:admmHG}}} &
\subfloat[Final reduce cell architecture]{\makebox[.5\linewidth][c]{\includegraphics[height = 0.9in]{images/admm/reducegraphup.png}\label{fig:admmRG}}}
\end{tabular}
\caption{Best trials of +ADMM+FIMT (a-d) and +ADMM+FIMT+CRB (e-h).}
\label{fig:admm_fimt}
\end{figure*}